\documentclass[conference]{IEEEtran}

\usepackage[dvipsnames]{color}
\usepackage{amsmath}
\usepackage{amsfonts}
\usepackage[amsmath]{ntheorem}
\usepackage{graphicx}
\usepackage[colorlinks=true, allcolors=blue]{hyperref}
\usepackage[nolist]{acronym}
\usepackage{blkarray}
\usepackage{amssymb}
\usepackage{babel,blindtext}
\usepackage[capitalize]{cleveref}
\usepackage{algorithm}
\usepackage{algpseudocode}
\usepackage{bbm}
\usepackage[normalem]{ulem}
\usepackage{mathtools}

\acrodef{ai}[AI]{Artificial Intelligence}
\acrodef{ml}[ML]{Machine Learning}
\acrodef{rl}[RL]{Reinforcement Learning}
\acrodef{snr}[SNR]{Signal to Noise Ratio}
\acrodef{sec}[SCE]{Semantic Channel Equalization}

\newcommand{\enc}{\lambda}
\newcommand{\dec}{\gamma}
\newcommand{\mt}{T}
\newcommand{\mct}{\mathcal{T}}
\newcommand{\q}{Q}
\newcommand{\obs}{o}
\newcommand{\sem}{x}
\newcommand{\act}{a}
\newcommand{\sspace}{\mathcal{X}}
\newcommand{\ospace}{\mathcal{O}}
\newcommand{\aspace}{\mathcal{A}}

\newcommand{\ls}{s}
\newcommand{\lt}{t}

\newcommand{\satom}{P}
\newcommand{\satomsource}[1]{P_{#1}^{\ls}}
\newcommand{\satomtarget}[1]{P_{#1}^{\lt}}
\newcommand{\infotransfer}[2]{\rho_{\satomsource{#1}\xrightarrow{}\satomtarget{#2}}} 
\newcommand{\pisem}{\pi_\text{sem}}
\newcommand{\pieff}{\pi_\text{eff}}
\newcommand{\quotes}[1]{``#1''}

\DeclareMathOperator*{\argmax}{arg\,max}

\newcommand{\titleheader}{This work has been accepted for publication in 2024 International Symposium on Wireless Communication Systems (ISWCS)}
\makeatletter
\def\ps@IEEEtitlepagestyle{
\def\@oddhead{\mbox{}\scriptsize \titleheader \rightmark \hfil }
}
\makeatother

\title{Soft Partitioning of Latent Space  for Semantic Channel Equalization}
\author{\IEEEauthorblockN{Tomás Hüttebräucker, Mohamed Sana, Emilio Calvanese Strinati}
\IEEEauthorblockA{CEA-Leti, Université Grenoble Alpes, F-38000 Grenoble, France\\
Email : \{tomas.huttebraucker; mohamed.sana; emilio.calvanese-strinati\}@cea.fr}}

\begin{document}
\maketitle
\begin{abstract}
Semantic channel equalization has emerged as a solution to address language mismatch in multi-user semantic communications. This approach aims to align the latent spaces of an encoder and a decoder which were not jointly trained and it relies on a partition of the semantic (latent) space into atoms based on the the semantic meaning. In this work we explore the role of the semantic space partition in scenarios where the task structure involves a one-to-many mapping between the semantic space and the action space. In such scenarios, partitioning based on hard inference results results in loss of information which degrades the equalization performance. We propose a soft criterion to derive the atoms of the partition which leverages the soft decoder's output and offers a more comprehensive understanding of the semantic space's structure. Through empirical validation, we demonstrate that soft partitioning yields a more descriptive and regular partition of the space, consequently enhancing the performance of the equalization algorithm.
\end{abstract}

\section{Introduction}
Semantic communication, as introduced by Weaver \cite{weaver1953recent} in its prelude to Shannon's seminal paper \cite{shannon1950mathematical}, is a paradigm where communication serves as a means to solve a given task rather than an end in itself. Semantic communication systems can significantly reduce overall network rate requirements by extracting and transmitting only task-relevant information from the data. Recently, they have been identified as a key enabler of future communication systems \cite{strinati20216g} \cite{kountoris2021} \cite{strinati2024goal}. The main driver behind the recent popularity of semantic communications is the success of \ac{ai}, particularly \ac{ml}, for automatic task solving. \ac{ml} can be utilized to learn a language (communication protocol) that enables effective communication and collaboration between connected agents . These communication protocols are well-suited for future technologies like smart cities and autonomous vehicles, where multiple intelligent agents communicate and collaborate to solve downstream tasks. Consequently, the design and development of such protocols are of paramount importance.

The development of \ac{ml} semantic protocols is an active area of research. Studies have shown that the bandwidth or energy consumption of wireless networks can be greatly reduced while still successfully completing downstream tasks without any performance loss \cite{bourtsoulatze2019deep} \cite{tung2021effective}. However, most literature assumes that the language between transmitter and receiver is shared as a result of a joint learning procedure, which might not hold in practical scenarios. In future networks, the set of agents participating in communication and collaboration to solve a given task will likely dynamically change depending on geographical factors and resource availability. In such cases, constantly re-learning a shared language proves to be a resource-intensive endeavor, which is infeasible in networks characterized by constrained energy and bandwidth resources. Conversely, when the language between agents is not shared, semantic mismatch arises, and task performance significantly drops \cite{sana2022learning}. To solve this, \ac{sec} was introduced \cite{sana2023semantic}. This framework models language as a partition of the communication space into multiple atoms, with each atom associated with a distinct semantic meaning. By doing so, \ac{sec} facilitates effective semantic translation among different languages with a low complexity algorithm. The efficacy of this approach in various domains, including image classification \cite{sana2023semantic} and reinforcement learning scenarios \cite{huttebraucker2024pragmatic} was extensively shown. While previous studies showcase the empirical effectiveness of the \ac{sec} framework, they do not delve into the crucial role of the semantics captured by the languages.

In this work, we aim to elucidate how different partitions of the latent space capture diverse semantic meanings and how these variations impact the equalization algorithm. Our primary contribution lies in introducing a novel methodology for partitioning the latent space, which captures richer semantic meaning and consequently enhances the performance of \ac{sec}.

\section{System Model}
\begin{figure}
    \centering
    \includegraphics[width=\linewidth]{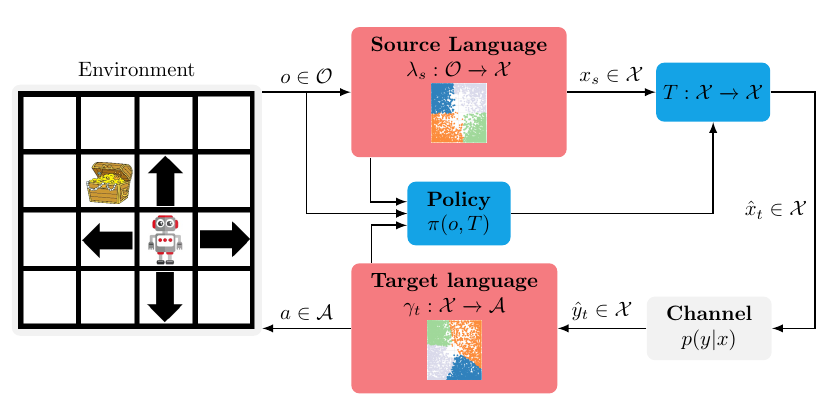}
    \caption{Proposed communication scenario shared with \cite{huttebraucker2024pragmatic}. A distributed control problem is explored, where the language of the encoder does not match the language of the decoder. Using a codebook of transformations and a selection policy, semantic channel equalization is performed. }
    \label{fig:system_model}
\end{figure}
Consider the distributed inference problem illustrated in \cref{fig:system_model}. Here, an encoder $\enc:\ospace\to\sspace$ transforms observations $\obs\in\ospace$ from the environment into a semantic representation $\sem\in\sspace$. This semantic representation $\sem$ encapsulates the relevant information present in $\obs$, which the decoder $\dec:\sspace\to\aspace$ interprets to select an action $\act\in\aspace$ (we will assume $\aspace$ to be discrete). In this study, we focus on the scenario where the communication between the encoder and the decoder is through a noisy channel over multiple time steps. It is worth noting that this general formulation includes image classification, which can be seen as a specialized case with only one time-step. Following the terminology introduced in \ac{sec} \cite{sana2023semantic}, we refer to $\enc$ as the \textbf{language generator} and to $\dec$ as the \textbf{language interpreter}. Furthermore, we denote $\ospace$, $\sspace$, and $\aspace$ as the observation space, semantic space, and action space, respectively and we denote $\mu_\obs$ as the probability distribution of observations. We assume that the language (communication strategy) between $\enc$ and $\dec$ is a result of a joint learning process. Through this process, the agents learn $\q(\act,\obs)$, an approximation of the \textit{true} action-value function $\q^*(\act,\obs)$, which is an indicator of how \quotes{good} it is to play action $\act$ when $\obs$ is observed. %We call 
Based on the learned action-value function $\q$, the agents make decisions following a \textit{greedy} policy:  $\dec(\enc(\obs))=\argmax_{\act\in\aspace}\q(\act,\obs)$.

\begin{figure}[h]
    \centering
    \includegraphics[width=\linewidth]{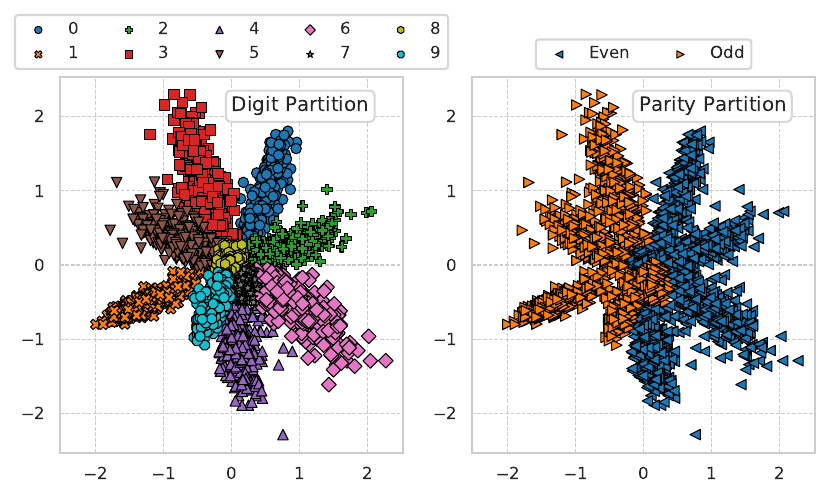}
    \caption{Two possible partitions for the latent space of a language generator. The generator was trained with the task of MNIST classification. Different ways to partition the semantic space depend on the criteria used (digit classification or digit parity classification)}
    \label{fig:mnist_partition}
\end{figure}

\subsection{Semantic space partitioning and language effectiveness}
A language generator $\enc$ defines a method for encoding information into the semantic space. When the generator operates effectively, it ensures that the task-relevant data features of each observation $\obs$ are appropriately encoded in the semantic representation $\sem = \enc(\obs)$. The resulting semantic space encapsulates all pertinent information in a structured manner, where observations with shared semantic characteristics are encoded similarly. This structure can be exploited to partition the space into multiple subspaces termed \textit{atoms}. Each atom corresponds to a distinct semantic meaning, with all semantic representations within an atom reflecting observations that share that particular meaning. We denote the set of chosen atoms as a \textit{partition} of the semantic space, represented as $\satom = \{\satom_0, \satom_1, \dots, \satom_N\}$, where $\satom_i$ denotes the $i$-th atom of the partition.

Different semantic space partitions can capture different levels of semantics. For example, in an image classification task, each atom of the semantic space can be associated with a label of the images, which is a high level description of semantics. However, if the encoder is descriptive enough, lower-level descriptions are also possible, and features such as the colors or shapes can be captured with the appropriate partitioning. As an example, see \cref{fig:mnist_partition} where two different partitions of the semantic space are shown for a generator trained to solve the MNIST classification task. Points in the latent (semantic) space are partitioned according to different criteria. One partition captures the digit information present in the semantic symbols while other possible partition captures the parity of this symbols. 

\subsection{Language mismatch in multi-user communication}
When the language generator and the language interpreter are not trained jointly, it is unlikely that they employ the same language and a semantic mismatch arises even if the training procedures follow the same architectures, data and objective functions \cite{moschella2022relative}. In \ac{sec} \cite{sana2023semantic}, the semantic mismatch arising between a language generator and a language interpreter was modeled as a misalignment of the atoms of their corresponding partitions. More precisely, when the source generator (transmitter) sends a message $\sem_s=\enc_s(\obs)$, it will not be interpreted correctly at the target interpreter (receiver) if it does not fall in the corresponding target atom of $\sem_t=\enc_t(\obs)$. We denote the source $(\enc_s,\dec_s,\sspace_s,\aspace_s,\ospace)$ and target $(\enc_t,\dec_t,\sspace_t,\aspace_t,\ospace)$ languages which were trained independently on the same observation space and we will explore the case where $\enc_s$ and $\dec_t$ communicate. 

\begin{figure*}[!t]
    \centering
    \includegraphics[width=.8\linewidth]{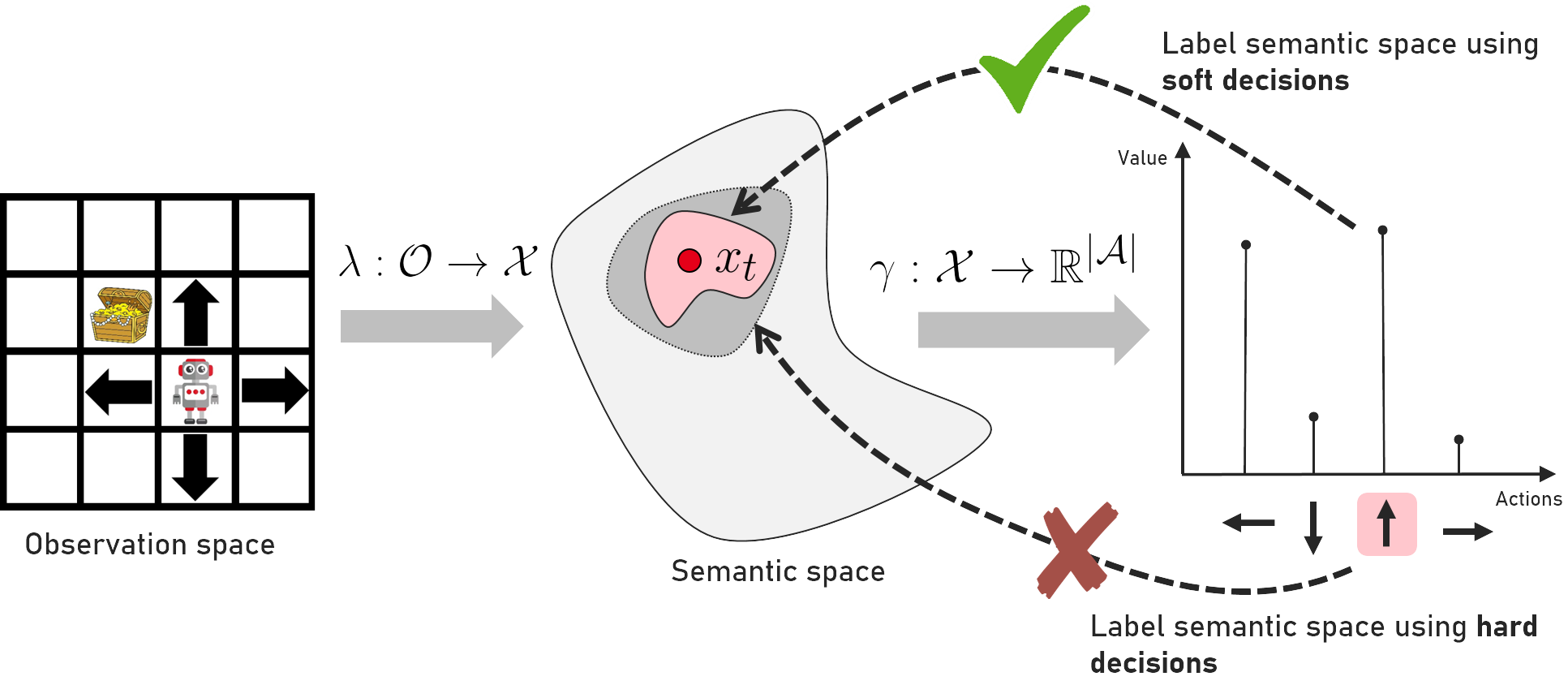}
    \caption{Different ways to partition the semantic space capture different semantics. When using the hard decision outcome to define the partition, the structure of the task and the relationship between actions is lost. When using the soft decision values, all the task-relevant information is exploited for the partition.}
    \label{fig:soft_parition}
\end{figure*}

\subsection{Compensating for language mismatch via Semantic Channel Equalization}
To deal with the language mismatch, the \ac{sec} algorithm leverages a codebook $\mct$ of linear transformations between individual atoms and a selection policy to operate the codebook. For a source $\satom^s = \{\satom_0^s, \satom_1^s, \dots, \satom_{J_s}^s\}$ and target $\satom^t = \{\satom_0^t, \satom_1^t, \dots, \satom_{J_t}^t\}$ partition of the semantic space, each transformation $\mt:\sspace_s\to\sspace_t\in\mct$ is learned using optimal transport and aims to maximize the transported volume for a given pair of source and target atoms:
\begin{equation}\label{eq:information_transfer}
\begin{split}
    \infotransfer{i}{j}(\mt) =   \frac{\mu_{\mt\enc_s}\left(\mt\left (\satomsource{i} \right )\cap \satomtarget{i}\right)}{\mu_{\mt\enc_s}\left(\satomsource{i}\right)}.
\end{split}
\end{equation}
Here $\mu_{\mt\enc_s}$ is the post-transformation distribution on the semantic space, which depends on the observation distribution $\mu_\obs$ and, if $\enc$ and $\mt$ are injective, can be written as $\mu_{\mt\enc_s}=\mu_\obs\circ\enc_s^{-1}\circ\mt^{-1}$. The codebook $\mct$ has as many transformations as the total pairs of source and target atoms, i.e. $|\mct|=J_s\cdot J_t$.

The operation policy selects a transformation $\mt$ from the codebook of transformation following 
\begin{equation}\label{eq:pisem}
    \pisem = 
    \argmax_{\mt\in \mct}\Bigg[ \sum_{i=1}^{J_s} \mu_{\enc_s}\left (\satomsource{i}|\obs \right)  \sum_{j\in\kappa(i)} \infotransfer{i}{j}(\mt) \Bigg ]
\end{equation}
where $\mu_{\enc_s}$ is the source language distribution (which can be written as $\mu_{\enc_s}=\mu_\obs\circ\enc_s^{-1}$ if $\enc_s$ is injective) on the semantic space and $\kappa(i)$ is a (problem dependent) mapping function between source and target atoms. The policy $\pisem$ aims to perfectly align source and target atoms according to their semantic meaning without regards to the downstream task performance. On \cite{huttebraucker2024pragmatic}, a new equalization policy which aims to maximize downstream task performance was proposed as
\begin{equation}\label{eq:pieff}
    \pieff =  \argmax_{\mt\in\mct}\Bigg[ \sum_{i=1}^{J_s} \mu_{\enc_s}\left (\satomsource{i}|\obs \right) \sum_{j=1}^{J_t} \infotransfer{i}{j}(\mt) \q_t(\act_j,\obs) \Bigg ].
\end{equation}
Where $\q_t(\act,\obs)$ is the target language's estimation of the true action-value function. Here, it is implicitly assumed that each target atom should correspond to a unique action, this is the assumption we challenge in this work. The policy $\pieff$ aims to maximize performance rather than perfect semantic alignment, which is not required to complete the task. We call $\pisem$ and $\pieff$ the semantic and effectiveness equalization policies respectively.

The effectiveness of \ac{sec} heavily relies on how the semantic space is partitioned. Essentially, this partitioning serves as a means of compressing the information intended for transmission. It groups distinct observations sharing the same semantic meaning into atoms. As \ac{sec} aligns these atoms, only the information captured by the partition is transmitted. Therefore, the selection of the language partition is a critical aspect of \ac{sec} as it determines the relevant information to be conveyed. If the chosen partition isn't suitable for the downstream task, the equalization process will likely fail. For instance, consider \cref{fig:mnist_partition}, showcasing two partitions of the semantic space generated by a MNIST classification model. If the objective is to classify digits, only the left partition captures the necessary semantics and aids the receiver effectively. Conversely, if the task involves classifying data parity, both partitions will convey the required semantics. However, opting for a more detailed partition comes with a trade-off—a more intricate equalization algorithm due to the increased number of atoms. This study aims to identify the optimal approach to partitioning the semantic space, considering the underlying task structure.

\section{The Impact of Semantic Space Partitioning on Equalization Performance}
\subsection{Hard partitioning}
To partition the semantic space, previous work considers hard partitioning \cite{sana2023semantic, huttebraucker2024pragmatic}. This approach defines an atom $\satom_i$ of a partition as the set of semantic symbols (i.e., states being mapped) in the semantic space that result in the same action $\act_i\in\aspace$: 
\begin{align}
       P_i = &\big\{\sem\in\sspace ~|~ \sem=\enc(\obs);\\ \nonumber
       &~\dec(x) = a_i=\argmax_{\act\in\aspace}\q(\act,\obs),~\forall \obs\in\ospace \big\} 
\end{align}
This approach is based on one assumption: there exists a one-to-one relationship between the atoms of the partition and the actions. Yet, in many control tasks, different possible actions may exist for a given observed state, thus leading to \textit{action ambiguities}. Action ambiguities can be detrimental to equalization performance when ignored. Indeed, when action ambiguity is present, hard partitioning assigns the semantic symbol $\sem$ to the output of the decision $\dec(\sem)$, ignoring its true semantic meaning. We show later that this approach leads to irregular atom shapes, which are hard to equalize. As an alternative, we propose a soft-value based partitioning, which leverages the action-values $\q(\cdot,\obs)\in\mathbb{R}^{|\aspace|}$.

\subsection{Proposed solution via soft-values based partitionning}
On \cref{fig:soft_parition} the main idea behind soft values semantic space partitioning is shown. Using the information present in the soft decision process of the interpreter allows for a more descriptive partition of the semantic space. Following the figure, if hard decisions were to be used for partitioning, the state will fall withing the semantic atom corresponding to the the estimated optimal action. However, there are two actions that lead to optimal behaviour. The learned $\q$ values scores capture this by assigning high value to these actions, where the difference between them is only due to noise in the learning process. Using the information provided by $\q$, the partition of the semantic space could be able to differentiate between states in which only one action is optimal and states where multiple actions are. Not only this is a better description of the semantic space, but also boosts the performance of the equalization algorithm, since, as we will show, the resulting atoms will be more regular.

We propose then to build the partition on the action-value space to build the atoms. For each observation $\obs\in\ospace$, we can represent the function $\q(\cdot,\obs)$ as a vector in the space $\mathbb{R}^{|\aspace|}$. Using standard clustering techniques, it is possible to divide the action-value space and translate this into the semantic space. More precisely, using a clustering algorithm $C:\mathbb{R}^{|\aspace|}\to \left\{ 0, 1, \ldots, n_c-1 \right\}$ which maps each point in the action value space $\mathbb{R}^{|\aspace|}$ into an index value indicating belonging to one of $n_c$ atoms, we can define a partition of the semantic space
\begin{equation}
    \satom = \{ \satom_0, \ldots, \satom_{n_c-1} \}
\end{equation}
where each atom is constructed following
\begin{equation}
    \satom_i = \left\{ \enc(\obs) | \obs \in \ospace \text{ and } C\left(\q(\cdot,\obs)\right) = i \right\}.
\end{equation}
The choice of the clustering algorithm is not simple and it is most likely problem dependent. In our work we choose to use the well known k-means algorithm \cite{mohiuddin2020kmeans}. This algorithm requires to define the number of atoms $n_c$ beforehand so we test multiple atom numbers in our experiments. While we are aware of the limitations of the algorithm with respect to convergence, the need to define the number of clusters beforehand and also the clustering criteria, we chose k-means both for its simplicity and popularity. Finding the optimal clustering criteria is beyond the scope of this paper, our objective is rather to show the influence of different latent space partition on \ac{sec}. 

\begin{figure}[!t]
    \centering
    \includegraphics[width=.8\linewidth]{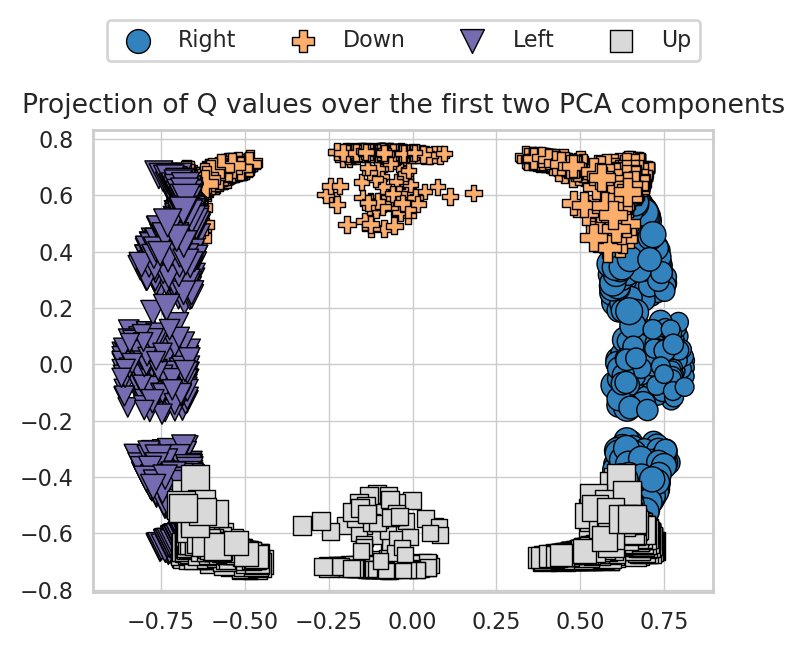}
    \caption{Projection of the action-value space of dimension $n_a=4$ into the first two data maximum variance directions for the source language. Each point corresponds to an observation. Colors are shown according to the action that maximized the value for each observation.}
    \label{fig:q_values_pca}
\end{figure}

\begin{figure*}[!t]
    \centering
    \includegraphics[width=\linewidth]{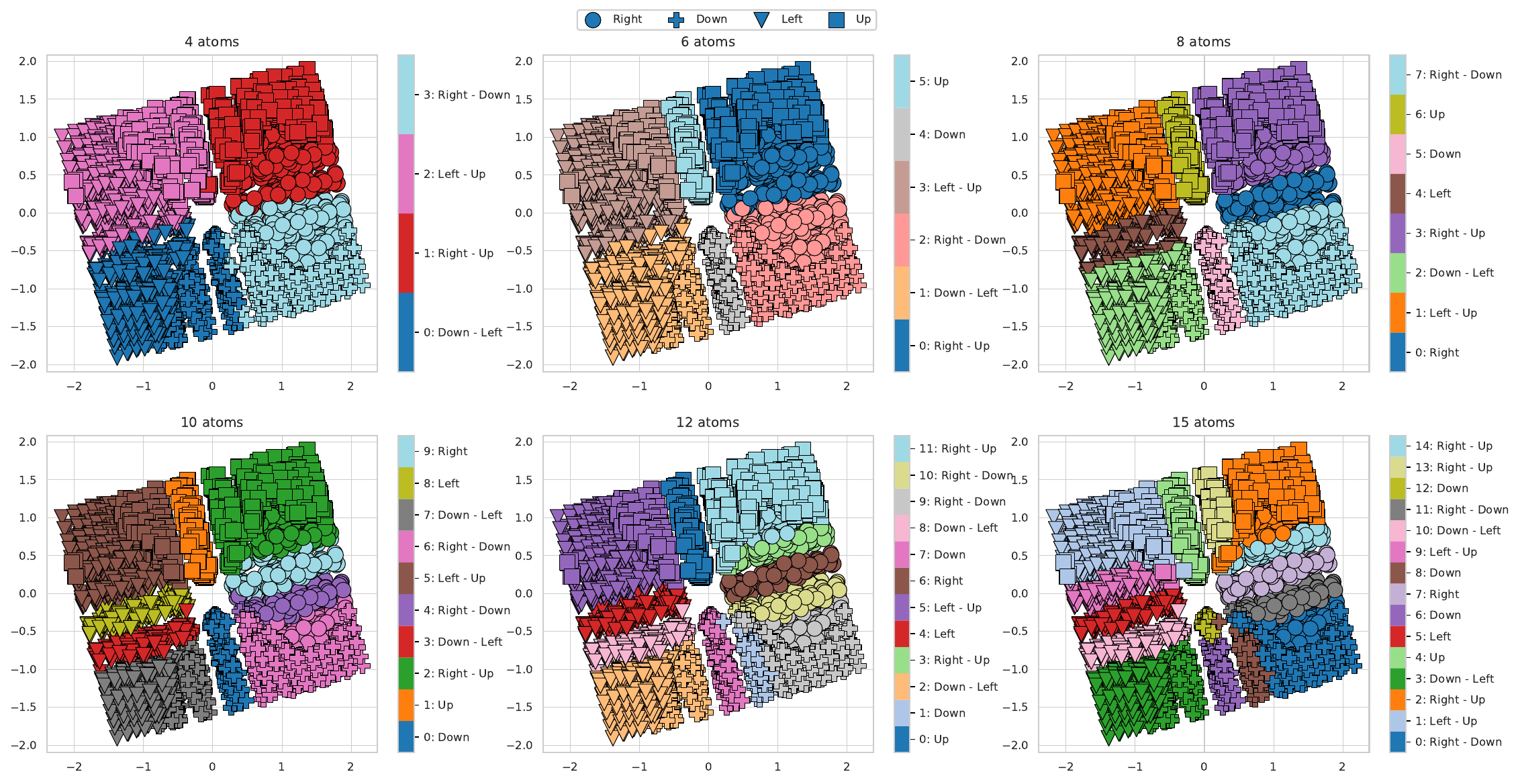}
    \caption{Different partitions of the semantic space using the k-means algorithm with varying number of atoms. The actions (hard partitioning) are visualized as different shapes for each partition. The color of the points corresponds to a given atom and the actions associated with each atom are shown next to it in the color legend. }
    \label{fig:task_clustering}
\end{figure*}

\section{Numerical results}
We evaluate the proposed system using a language generated through \ac{rl} techniques, aiming to address the environment depicted in \cref{fig:system_model}. In this case, the observation space $\ospace$ represents the state of a grid world featuring an agent and a treasure. The encoder $\enc$ maps each observation into $\sspace=\mathbb{R}^2$. During training, the average output power of the encoder is standardized to 1. This is achieved through the implementation of a rolling mean normalization technique, where a normalization constant $\tau$ is computed as $\tau_i = \eta \cdot \tau_{i-1} + (1-\eta) ||\sem^*_i||^2$. Here, $\sem^*_i\in\sspace$ denotes the non-normalized semantic symbol ($\sem_i=\sem_i^*/\tau_i$) chosen at training step $i$, with $\eta=0.1$ representing the momentum value. During testing, the value of $\tau$ is fixed as the final value obtained during training. Subsequently, the decoder $\dec$ processes the noisy version of the transmitted symbol and selects an action from $\aspace=\{\text{right, down, left, up}\}$, which the agent executes. An episode concludes either when the agent reaches the treasure or when the maximum number of steps (150) is attained. We operate under the assumption that the encoder decoder pairs $(\enc_s,\dec_s)$ and $(\enc_t,\dec_t)$ are provided and have each undergone a joint training in a centralized manner utilizing \ac{rl} techniques, all while adhering to the same reward signal and a \ac{snr} of 5 dB. For training purposes, we employ Deep Q-Learning (DQN) \cite{mnih2015human}, which is well-suited for discrete action spaces. Additionally, we set $\sspace_s=\sspace_t=\sspace$ and $\aspace
_s=\aspace_t=\aspace$.

\subsection{Semantic space partitioning}
We first explore the possible semantic space partitions for the learned languages. We only show the results for the source language for lack of space. We first show that the information included in the action-value space is more descriptive of the task than the hard-decision clustering. On \cref{fig:q_values_pca} the projection of the four dimensional action-value space for the source language is shown. It is clear that, even if the total amount of actions is four, the action-value space hints that different partitions are possible. It is easy to see at least eight clusters, four corresponding to unique actions (originated from states where only one action is optimal) and four others corresponding to pairs of actions (originated from states where more than one action is optimal, as such shown on \cref{fig:soft_parition}). 

On \cref{fig:task_clustering} we show the resulting partitions on the semantic space when using k-means on the action-value space for multiple choices of number of atoms $n_c$. In general, we can observe that the resulting atoms are more regular in shape compared to the hard partitioning (which is indicated by the shape of the plotted points). Different partitions capture different semantic descriptions of the task. In particular, for $n_c=8$ the eight atoms of the semantic space correspond to the observed clusters in the action-value space on \cref{fig:q_values_pca}. However, for four and six atoms, the semantic meaning for some individual actions is lost, as we will show next, this will be detrimental for the performance of the equalization. 

\subsection{Performance of proposed solution}
Leveraging the partitions of the semantic space shown on \cref{fig:task_clustering}, we implement both equalization policies $\pisem$ and $\pieff$ introduced by the \ac{sec} framework as described on \cref{eq:pisem} and \cref{eq:pieff} respectively. When using k-means based soft partitioning, we replace $\mu_{\enc_s}$ by the normalized inverse of the distances to the k-means centers and $\kappa(i)$ as the index of the target atom whose center in the action-value space lies closest to the one from $\satomsource{i}$. To compute the action value of each target atom (equivalent to $\q_t$ in \cref{eq:pieff}) we use the average Q-value for the actions in it, i.e. 
\begin{equation}
    \q_t(\satomtarget{i},\obs)=\frac{1}{|\satomtarget{i}|}\sum_{\sem\in\satomtarget{i}}\q_t(\gamma_t(\sem_j),\obs) .
\end{equation}

\begin{figure}[!t]
    \centering
    \includegraphics[width=.95\linewidth]{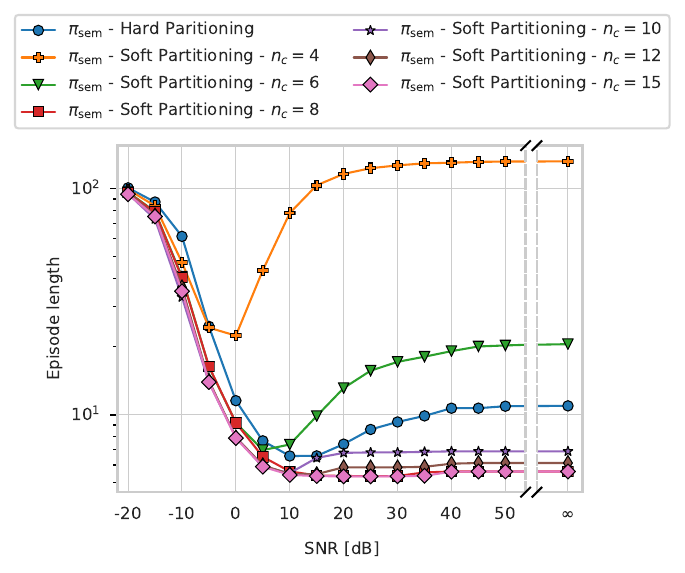}
    \caption{Performance of the policy $\pisem$ as a function of SNR for hard partitioning and soft partitioning with different numbers of atoms $n_c$.}
    \label{fig:pisem}
\end{figure}

\begin{figure}[!t]
    \centering
    \includegraphics[width=.95\linewidth]{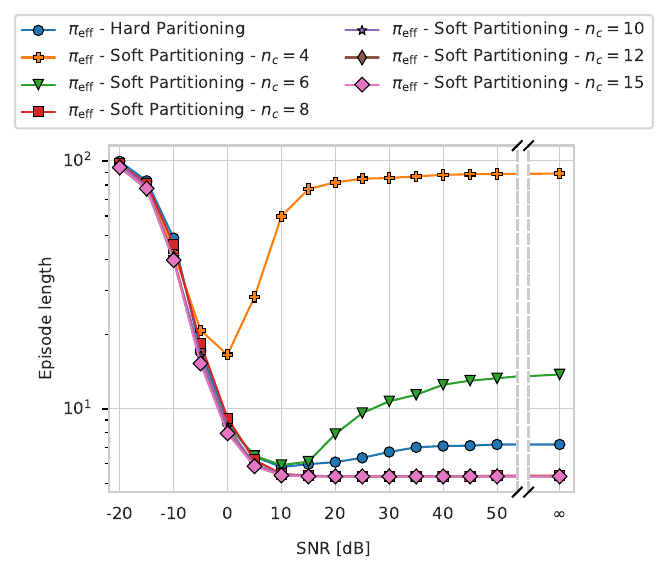}
    \caption{Performance of the policy $\pieff$ as a function of SNR for hard partitioning and soft partitioning with different numbers of atoms $n_c$.}
    \label{fig:pieff}
\end{figure}

The results for $\pisem$ are shown on \cref{fig:pisem} and for $\pieff$ on \cref{fig:pieff}. The performance of both policies depends on the partition of the semantic space and we note that soft partitioning is not always beneficial. For example, when using soft partitioning with four and six atoms, the performance of the equalization is worse than when using the four action hard partitioning. To understand why, it suffices to look at the resulting partitions on \cref{fig:task_clustering}. For the case of four and six atoms, all unique actions do not have a corresponding atom, as shown by the atom labels. For four atoms, no unique action is captured, and for six atoms, unique actions \quotes{left} and \quotes{right} are not captured. This degrades the performance, as the equalization algorithm can not transmit the message of taking only a singular action, which in the particular task is necessary to reach the goal. However, when the soft partition allows to capture all the relevant information of the task, the performance is greatly improved. From our results, we can conclude that the best number of atoms for our problem is eight, since it is the smallest number of atoms for which we obtain the best performance. This observation reinforces our intuition on the optimal number of atoms, in which each atom should be associated to all singular actions and all possible action ambiguities, which, for this particular example, gives us a total of eight atoms. 

\subsection{On the descriptiveness of soft partitioning for multi-task equalization}
Soft partitioning offers a more nuanced depiction of the semantic space, enabling a richer understanding of semantics that can be beneficial for multi-task equalization. For instance, consider a scenario where the decoder is expanded to accommodate eight actions, including diagonal moves, instead of the original four horizontal and vertical ones. In this case, hard partitioning may hinder the encoder's ability to transmit semantic meanings associated with diagonal actions. On the other hand, soft partitioning the semantic space into eight atoms (as illustrated in \cref{fig:task_clustering}) could empower the decoder to interpret atoms containing information about double actions as diagonal moves. This straightforward example underscores the potential of soft clustering for multi-task equalization. Further exploration of this aspect is left for future research.

\section{Conclusions}
Motivated by the recent advancement on Semantic Channel Equalization, in this work we address the role of the semantic space partitioning on the equalization performance. We first justify why partitioning the semantic space according to output actions (hard partitioning) is sub-optimal in cases where there are action ambiguities, i.e. multiple optimal actions. To address this problem, we propose to use soft partitioning, which leverages the estimates action-values to  define the semantic atoms. We show that, using soft partitioning, the resulting partition of the space is more regular, which improves the equalization performance. Moreover, we show that richer semantics of the problem can be captured by soft partitioning, which is a promising result that can be applied to multi tasking equalization. 

\section*{Acknowledgements}
The present work was supported by the EU Horizon 2020 Marie Skłodowska-Curie ITN Greenedge (GA No. 953775), by ``6G-GOALS", an EU-funded project, and by the French project funded by the program \quotes{PEPR Networks of the Future} of France 2030.

\bibliographystyle{ieeetr}
\bibliography{bib}

\end{document}